\documentclass{article}

 \PassOptionsToPackage{numbers,compress}{natbib}

\usepackage[preprint]{neurips_2025}

\usepackage[utf8]{inputenc} 
\usepackage[T1]{fontenc}    
\usepackage{hyperref}       
\usepackage{url}            
\usepackage{booktabs}       
\usepackage{amsfonts}       
\usepackage{nicefrac}       
\usepackage{microtype}      
\usepackage{xcolor}         
\usepackage{enumitem}       
\usepackage{makecell}
\usepackage{multirow}
\usepackage{graphicx}
\usepackage{amsmath}
\usepackage{array}
\usepackage{subcaption}
\usepackage{caption}

\title{Response Uncertainty and Probe Modeling: Two Sides of the Same Coin in LLM Interpretability?}

\author{
  Yongjie Wang, Yibo Wang, Xin Zhou, Zhiqi Shen \\ 
  Nanyang Technological University, Singapore\\
  \texttt{\{yongjie.wang, xin.zhou, zqshen\}@ntu.edu.sg} \\
  \texttt{yibo003@e.ntu.edu.sg}
}

\begin{document}

\maketitle

\begin{abstract}
Probing techniques have shown promise in revealing how LLMs encode human-interpretable concepts, particularly when applied to curated datasets. However, the factors governing a dataset's suitability for effective probe training are not well-understood. 
This study hypothesizes that probe performance on such datasets reflects characteristics of both the LLM's generated responses and its internal feature space. Through quantitative analysis of probe performance and LLM response uncertainty across a series of tasks, we find a strong correlation: improved probe performance consistently corresponds to a reduction in response uncertainty, and vice versa.
Subsequently, we delve deeper into this correlation through the lens of feature importance analysis. Our findings indicate that high LLM response variance is associated with a larger set of important features, which poses a greater challenge for probe models and often results in diminished performance. Moreover, leveraging the insights from response uncertainty analysis, we are able to identify concrete examples where LLM representations align with human knowledge across diverse domains, offering additional evidence of interpretable reasoning in LLMs.
\end{abstract}

\section{Introduction}
Large Language Models (LLMs) have demonstrated remarkable performance across a wide range of tasks, including question answering \cite{edge2024local}, behavioral agency \cite{park2023generative,wang2024voyager}, and clinical assistance \cite{qiu2024llm,thirunavukarasu2023large}. Despite these impressive capabilities, some fundamental questions persist: how next-token prediction training yields such broad and functionally useful behaviors, and more specifically, whether LLMs develop a genuine understanding of the physical world \cite{gurnee2024language,engels2025not}, or merely capture superficial statistical correlations \cite{bender-koller-2020-climbing}. To understand LLM internal mechanism, researchers propose various explanatory approaches \cite{singh2024rethinking,sharma2024the} for enhancing model trustworthiness, particularly in high-stakes applications like healthcare  \cite{qiu2024llm,thirunavukarasu2023large} or legal decision-making \cite{cheong2024not,colombo2024saullmb}. Among these, probing-based techniques are widely employed.

Probing-based explanations \cite{alain2017understanding,kim2018interpretability,belinkov-2022-probing,gurnee2024language,mikolov2013linguistic} aim to train a simple auxiliary model (known as probe) that predicts concept attributes (e.g., syntax, toxicity) from representations of a frozen model, thereby elucidating how concepts are encoded within its latent space. Specifically, Testing with Concept Activation Vectors (TCAV) \cite{kim2018interpretability} provides quantitative evidence that CNN models capture human-interpretable concepts (e.g., striped patterns). Recent studies \cite{patel2022mapping,abdou2021can} demonstrate that LLM representations exhibit significant alignment with human perceptual and conceptual systems in color and spatial domains. \citet{nanda-etal-2023-emergent} demonstrate that OthelloGPT \cite{li2023emergent} learns representations of board turn-taking states that could be probed linearly. \citet{gurnee2024language} reveal that LLMs encode temporal and spatial representations across multiple scales, whereas \citet{engels2025not} show non-linear embedding structures for cyclical temporal concepts (e.g., months/years), challenging the linear representation hypothesis posited by \cite{mikolov-etal-2013-linguistic}.

Building on the success of probing in extracting encoded concepts from LLM internal representations, and recognizing that LLM responses also originate from these representations, we hypothesize that a relationship exists between LLM responses and probing efficacy.
Validating this hypothesis serves to bridge the gap between two previously isolated fields: LLM probing and response uncertainty. This connection holds significant practical implications; for instance, it could streamline the identification of LLM-encoded concepts by mitigating the need for laborious data selection and annotation, as well as trial-and-error processes in conventional probing methods. 

To verify the aforementioned hypothesis, we conduct extensive experiments investigating the interplay between probe performance and the uncertainty observed in LLM generative responses.
Specifically, for each sample in a dataset (e.g., historical figure) in \cite{gurnee2024language}, we query the LLM multiple times with an identical prompt (e.g., ``Please return the death year of George Washington'') to collect a set of responses. We then compute the response uncertainty for each sample and sort all samples in descending order of this uncertainty. Subsequently, we partition the sorted dataset into several bins and train a separate probe model for each. Finally, we report the correlation between the average response uncertainty and probe performance across these bins. 
Our results indicate a strong correlation: when an LLM exhibits lower response uncertainty (i.e., less variability in its outputs), its corresponding internal representations can be more accurately modeled by a linear probe.

Moreover, we delve deeper into this observed phenomenon through the lens of feature attribution.
Specifically, we adopt AttnLRP (Attention-aware layer-wise relevance propagation) \cite{10.5555/3692070.3692076}, an adaptation of the LRP framework \cite{bach2015pixel} for transformer models, to analyze which features contribute most significantly to the LLM’s response. For latent representations at a specific LLM layer, our experiments reveal that responses with higher uncertainty tend to distribute relevance across a larger number of features compared to those with lower uncertainty, where relevance is more concentrated. This broader distribution of important features poses a significant obstacle for downstream probe models attempting to achieve a good fit, as supported by our theoretical analysis. 
From our analysis, we establish a clear link between response uncertainty and probe performance, which appear to be two sides of the same coin reflecting the nature of the LLM's internal representations.
Additionally, informed by these findings, we identify specific instances—such as hours, birth years of historical figures, and brand names—where the LLM representations exhibit strong consistency, aligning with human domain knowledge.

Our key contributions are summarized as follows:
\begin{itemize}[topsep=1pt,leftmargin=10pt]
    \item We conducted extensive experiments on six time and space datasets across six public LLMs, demonstrating a strong correlation between LLM response uncertainty and probe performance. 
    \item We provide a mechanistic explanation for this correlation using feature attribution, demonstrating that increased response uncertainty leads to relevance signals distributed across a greater number of features, thus hindering probe model performance. This empirical finding is further supported by our theoretical analysis.
    \item Our analysis of uncertainty patterns enables the identification of examples where LLM embeddings exhibit significantly consistent with human domain knowledge, yielding more interpretable insights into the model’s internal representations.
\end{itemize}

\section{Preliminaries}

\textbf{Response uncertainty:} LLMs often suffer from hallucination issues \cite{maynez-etal-2020-faithfulness,10.1145/3571730}, producing non-factual or fabricated responses that deviate from grounded knowledge. A key observation is that the model's confidence typically aligns with the correctness and consistency of its outputs, while low-confidence responses (high uncertainty) are more likely to be arbitrary or hallucinatory.
Leveraging this insight, response uncertainty has emerged as a valuable signal for detecting potential hallucinations. This uncertainty is commonly quantified by analyzing the variation across multiple response trials and measuring their semantic dispersion using metrics such as response log-likelihood \cite{kadavath2022language}, semantic entropy \cite{farquhar2024detecting,kuhn2023semantic,kossen2025semantic}, and other methods \cite{dwaracherla2023ensembles,pmlr-v235-johnson24a,wang2023gaussian}.

\textbf{LLM probing:} Let an LLM $f$ decompose into two modules $ f = \varphi \circ \phi $, where $ \phi $ maps a textual input $x$ to a latent representation $\phi(x) \in \mathbb{R}^d$, and $\varphi$ maps this latent representation $ \phi(x) $ to the final generation $ \varphi (\phi(x))$. A probing method \cite{alain2017understanding,belinkov-2022-probing,10.1145/3639372} aims to train a simple probe that predicts the presence or strength of a concept from the LLM latent representation $\phi(x)$. 
To formally define the probing setup, we consider training a probe model $g_{\theta}$, parameterized by $\theta$, to predict a target variable $Y$ from LLM representations $\mathcal{D}$. Here, $\mathcal{D} \in \mathbb{R}^{n\times d}$ represents the latent features for $n$ samples, and $Y$ indicates the presence or strength of a concept, taking either discrete \cite{kim2018interpretability,alain2017understanding,nanda-etal-2023-emergent} or continuous \cite{gurnee2024language} values. The probe is trained by minimizing a loss function with respect to its parameters $\theta$:
\begin{align}
L(\theta) = \frac{1}{n} \sum_{i=1}^{n} \ell\left(y_i, g_\theta(D_i)\right) + \lambda ||\theta||_2^2,
\end{align}
where $\ell(\cdot)$ is a loss function (e.g., cross entropy or mean square error) measuring the difference between the probe's prediction for the $i$-th sample and its ground truth $y_i$. High probe performance indicates that the LLM more effectively encodes information relevant to the target concept.

Although extensive studies have investigated either probing approaches or response uncertainty, these research areas are typically studied in isolation.
A critical gap persists in understanding the relationship between how well an LLM's internal representations can be probed for a concept (probe performance) and the uncertainty associated with its generated responses related to that concept. Elucidating this relationship is crucial: it could provide novel insights for diagnosing LLM hallucinations and enable more efficient probing strategies by indicating when the model is unlikely to have effectively encoded the target concept, thereby reducing the need for unnecessary data labeling.

Our idea is motivated by failure cases encountered when probing LLM embeddings for predicting product prices in the Amazon Fashion dataset \cite{hou2024bridging}. In this task, the probe model was unable to learn any meaningful pattern from the LLM’s embeddings, indicating low probe performance. Concurrently, when we directly queried GPT-4 for product prices, it produced highly inconsistent answers, suggesting high response uncertainty. 
This contrasts sharply with tasks known to be easily probeable \cite{engels2025not}, where GPT-4 reliably provides correct answers. These observations lead us to hypothesize that \textit{there exists a correlation between LLM response uncertainty and probe performance}.

\section{The interplay between probe performance and response uncertainty}
\label{sec:methodology}

\subsection{Methodology}

\begin{figure}
    \centering
    \includegraphics[width=\linewidth]{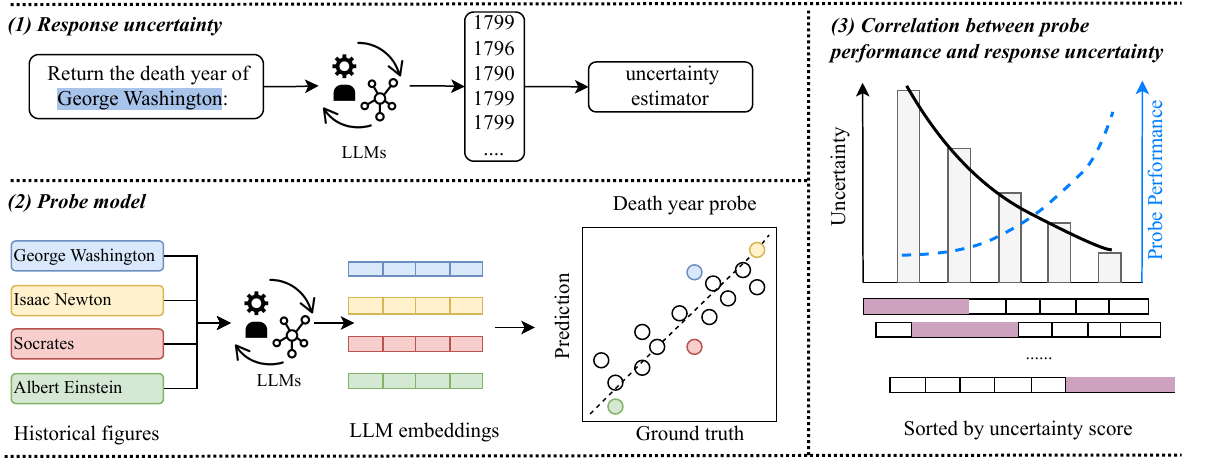}
    \caption{Framework for correlating probe performance with response uncertainty.}
    \label{fig:framework}
\end{figure}

To verify our hypothesis, we conducted experiments on six datasets from \cite{gurnee2024language}, spanning the domains of history, arts, news, and geography—each involving either time or space concept. For each sample in the dataset, the entity name and additional attributes—such as associated locations, death year, and release date—are recorded. For example, in the historical figures dataset, a sample such as ``George Washington'' includes attributes like birth year, death year, and other demographic information. \citet{gurnee2024language} train a separate regression probe to predict the temporal or space concept from the LLM embedding. More detailed description of six datasets is introduced in Appendix section \ref{apd:dataset}. 

The workflow of our method is demonstrated in Figure \ref{fig:framework}. First, for each sample in these datasets, we query an instruction-tuned LLM multiple times (typically 20) with the same prompt to predict the target concept—for example, ``\textit{Return the death year of George Washington}.'' Since LLMs could generate varying responses—such as ``1799'' or ``1796''—we use uncertainty measures to estimate the variability in response behavior. Unlike popular approaches that use entropy to measure uncertainty over discrete predictions \cite{farquhar2024detecting,kuhn2023semantic,kossen2025semantic}, our target concept are continuous values. This necessitates using variance as our default metric for response uncertainty \cite{bülte2025axiomatic}. Additionally, we also conduct experiment with entropy as an uncertainty estimator on temporal concepts.

Instead of training a single probe over entire samples for each dataset \cite{gurnee2024language}, we sort all samples in decreasing order of response uncertainty. We then divide the sorted dataset into several segments using a sliding window strategy, allowing overlap between consecutive segments. For each segment, we first report the average response uncertainty over all samples within it, and then train a separate probe to assess its performance. Finally, we report the correlation coefficient (e.g., \textit{Kendall and Spearman rank correlation}) between probe performance and average response uncertainty across multiple segments, as shown in Figure~\ref{fig:framework}.

For each segment of samples, we follow the methodology of \citet{gurnee2024language} to train the probe model. In particular, each data sample is fed into the same instruction-tuned LLM, and the hidden activations of a target layer are extracted as the data representation. We use the representation of the last token as the sample’s overall representation, following \citet{gurnee2024language}. Given a segment with $n$ samples, we obtain a representation dataset $\{D_l, Y\}$ for a target layer $l$, where $Y$ contains the ground truth of each sample-such as the death year, location, and $D_l$ is the representation set. We split the segment into training and testing sets, and train a linear regression model $\hat{Y} = D_l W$ by optimizing the following objective over the training set,
\begin{align}
    \hat{W} = \underset{W}{\arg\min} ||Y-D_lW||_2^2 + \lambda ||W||_2^2,
\end{align}
where $\lambda$ is the regularization term. We report standard regression metrics, including both the \textit{$R^2$ score and the Spearman rank correlation} on the test set, to evaluate probe performance of each segment. To stabilize probe performance, we randomly split each segment five times using different random seeds and report the average performance across the five trials. 

\subsection{Implementation details}
All of our experiments are conducted on a series of instruction-tuned LLMs, ranging from 8 billion to 72 billion parameters, spanning multiple LLM families including Llama-3.1 (8B, 70B), Qwen 2.5 (14B, 72B) \cite{yang2024qwen2}, Mistral-Small (24B), and Gemma-2 (27B) \cite{gemma_2025} series. As reported by \citet{gurnee2024language}, probe performance typically plateaus after the middle layers of LLMs, and our reproduced probe experiments also support this conclusion. By default, we use the activations of last token from the deeper layer (after the residual connection) as the sample representation. Since probe performance stabilizes after the middle layers, the choice of target layer does not affect our conclusions as long as it is no shallower than the middle layer.

For larger LLMs with higher-dimensional representations, we use a larger window size that contains more samples to mitigate overfitting. Additional details are provided in Appendix~\ref{apd:implementation}, including the window size and stride used for segment division, probe training configurations, and the target layer selected for each LLM.

\section{Experiment results}

\subsection{Existence of high correlation}

\begin{figure}[tbp]
  \centering
  \includegraphics[width=\linewidth]{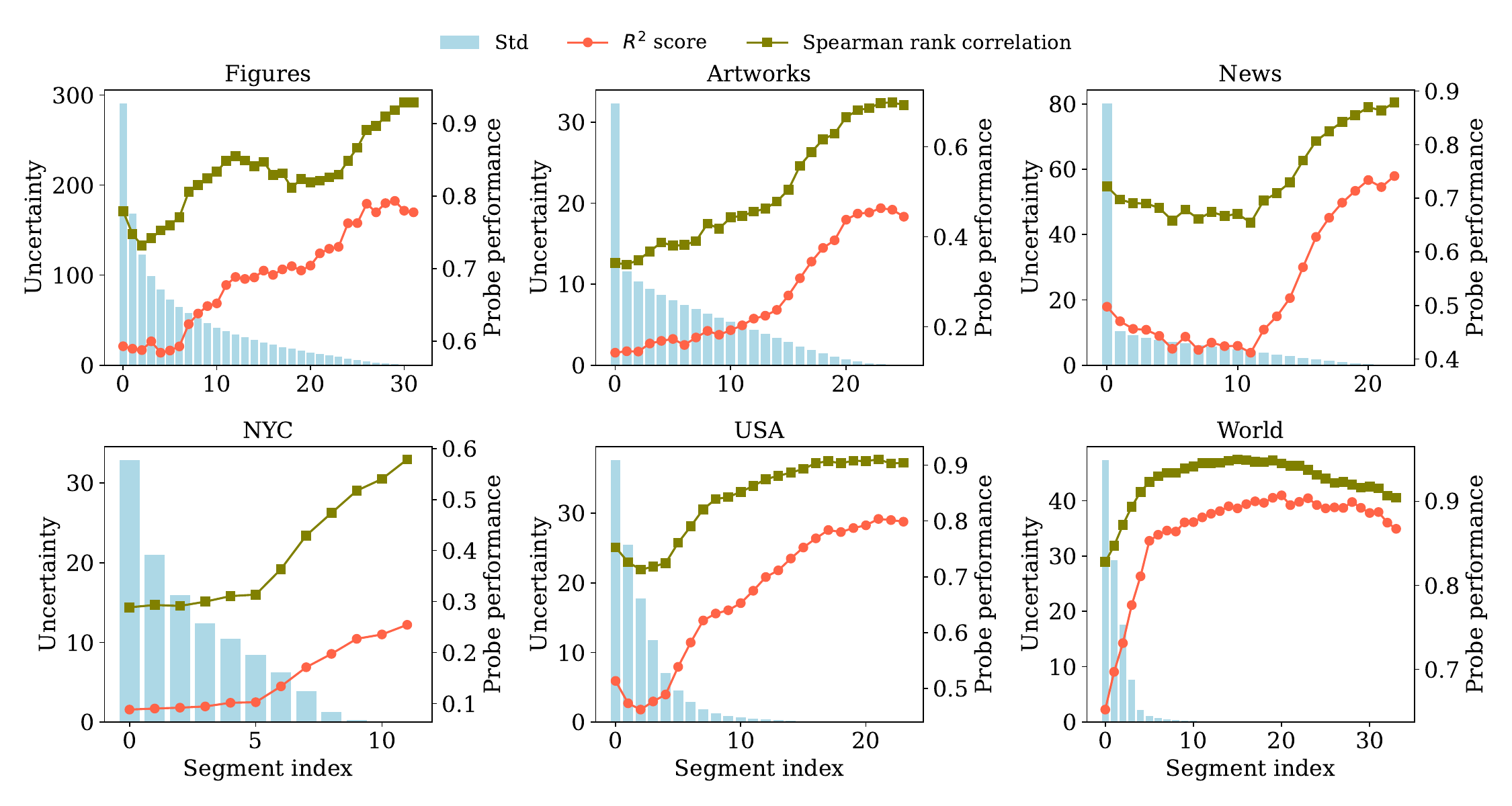}
  \caption{Trend between response uncertainty and probe performance on the Llama 3.1 (8B) model over all six datasets.}
  \label{fig:main_fig}

\begin{minipage}{0.65\textwidth}
  \centering 
 \includegraphics[width=\linewidth]{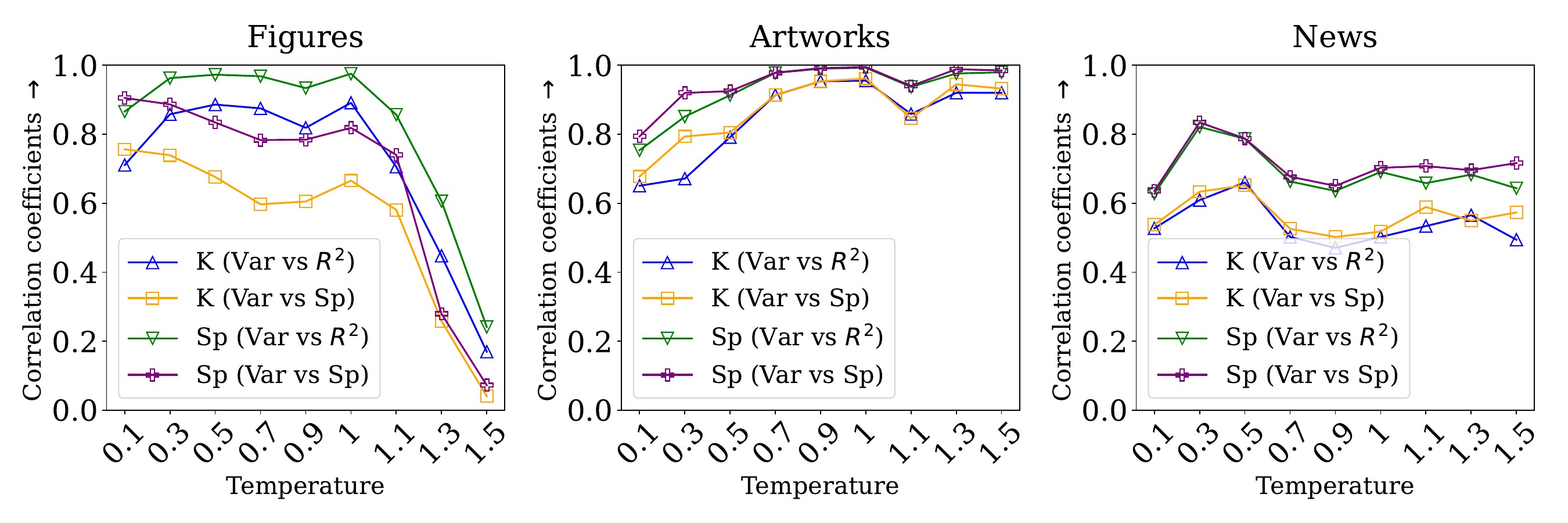}
  \caption{The correlation analysis with different temperatures in generation for Llama 3.1 (8B). `Var', `$K$', and `Sp' denotes variance (uncertainty estimator), Kendall and Spearman rank correlation coefficients. Here, we use absolute value for good visualization.}
  \label{fig:temperature}
\end{minipage}%
\hfill%
\begin{minipage}{0.32\textwidth}
  \centering
  \includegraphics[width=0.84\linewidth]{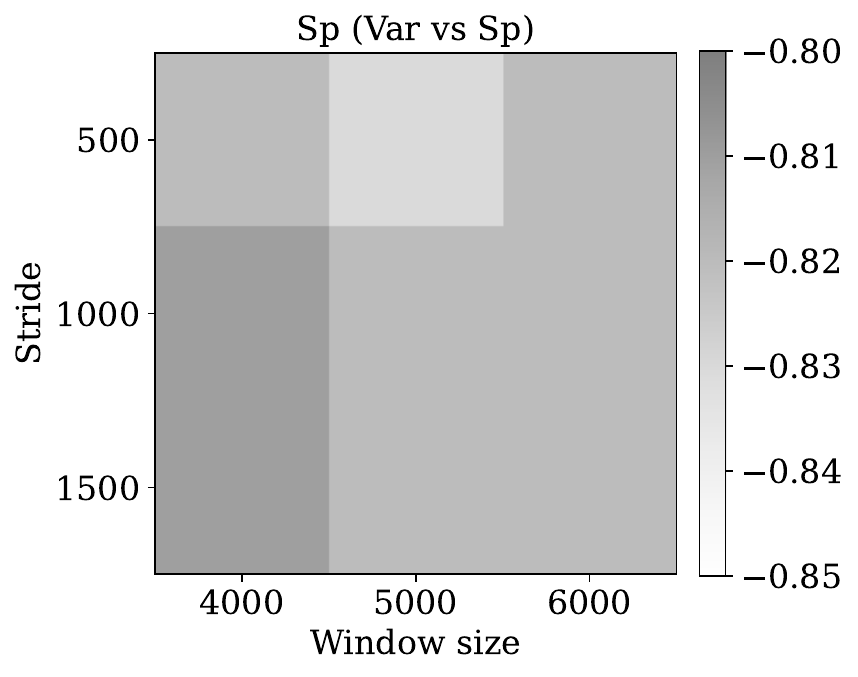}
  \caption{Correlation coefficient vs sliding window parameters, Llama 3.1 (8B) on Figures.}
  \label{fig:sliding_window}
\end{minipage}
\vspace{-0.5cm}
\end{figure}

In this experiment, we investigate the correlation between LLM response uncertainty and probe performance across six datasets using six public LLMs. As shown in Figure~\ref{fig:main_fig}, probe performance generally increases with decreased response uncertainty. Despite minor fluctuations, a strong negative correlation is observed. We report the correlation scores in Table~\ref{tab:main_results}, where both Kendall and Spearman rank correlation coefficients typically fall below $-0.5$, often approaching $-1$, indicating a strong negative correlation between these two factors. Exceptions include the USA dataset for larger models such as Llama 3.1 70B and Qwen 2.5 72B, where the LLMs produce highly consistent responses across most samples, making it difficult to validate our hypothesis. Similarly, on the World dataset, LLMs exhibit very low response uncertainty for many samples, thus, we report the correlation using only the top 20,000 most uncertain samples.

When response uncertainty is measured with entropy on the first three temporal datasets, the same conclusion still holds, as shown in the Appendix section \ref{apd:entropy_estimator}. (Note that entropy is not suited for location predictions, such as latitude and longitude, where even small numerical differences can correspond to large spatial differences on the map.) One interesting question is whether the correlation is linear. In our experiment (Appendix section~\ref{apd:linear}), we find that linearity depends on the choice of uncertainty estimator. Since entropy and variance operate on different scales depending on the numerical range of the ground truth, their behavior varies. When using entropy as the uncertainty measure, the Pearson correlation is strongly negative, whereas it is relatively low when using variance. We leave a deeper investigation into the conditions under which linearity holds in general cases to future work.

\begin{table}[tb]
    \centering
    \caption{Correlation coefficients between probe performance and LLM response uncertainty on Llama 3.1 (8B, 70B), Qwen 2.5 (14B, 72B), Mistral-Small (24B), and Gemma-2 (27B). Symbol `\textemdash' indicates uniformly low uncertainty over almost all samples; \underline{underscored} values are computed on the top 20,000 most uncertain samples. }
    \label{tab:main_results}
    \scalebox{0.9}{
    \begin{tabular}{l| cc | cc | cc | cc | cc | cc  cc}
    \toprule
    \multirow{2}{*}{Dataset} & \multicolumn{2}{c|}{LLama 8B} & \multicolumn{2}{c|}{LLama 70B} & \multicolumn{2}{c|}{Qwen 14B} & \multicolumn{2}{c|}{Qwen 72B} & \multicolumn{2}{c|}{Mistral 24B} & \multicolumn{2}{c}{Gemma 27B} \\ \cmidrule{2-13}
     & K & Sp & K & Sp & K & Sp &  K & Sp & K & Sp & K & Sp \\ \midrule
    \multicolumn{13}{c}{\textit{Correlation coefficient: response uncertainty (\textbf{variance}) vs probe performance (\textbf{$R^2$})}. } \\ \midrule
    Figures  & -0.89 & -0.98 & -0.76 & -0.88 & -0.71 & -0.83 & -0.79 & -0.92  & -0.86 & -0.95 & -0.96 & -0.99 \\
    Artworks & -0.94 & -0.99 & -0.97 & -0.93 & -0.81 & -0.93 & -0.79 & -0.94  & -0.90 & -0.97  & -0.89 & -0.97   \\
    News  & -0.57 & -0.71 & -0.92 & -0.98 & -0.45 & -0.67 & -0.41  & -0.68 & -0.78 & -0.90 & -0.93& -0.99 \\
    NYC  & -0.89 & -0.96 & -0.80 & -0.90 & -0.88 & -0.96 & -0.96 & -0.99  & -0.97 & -0.99 & -0.75 & -0.88  \\
    USA  & -0.91 & -0.98 & \textemdash & \textemdash & -0.86 & -0.94 & \textemdash & \textemdash  & -0.72 & -0.87 & -0.91 & -0.96\\
    World   &\underline{-0.80} & \underline{-0.92} & \underline{-0.89} & \underline{-0.91} & \underline{-0.69}  & \underline{-0.74} & \underline{-0.89} & \underline{-0.97} & \underline{-0.92} & \underline{-0.98} & \underline{-0.96} & \underline{-0.99} \\  \midrule
    \multicolumn{13}{c}{\textit{Correlation coefficient: response uncertainty (\textbf{variance}) vs probe performance (\textbf{Spearman rank correlation})}.} \\ \midrule
    Figures  & -0.67 & -0.82 & -0.77 & -0.90 & -0.82 & -0.95 & -0.55 &  -0.72 & -0.86 & -0.96 & -0.89 & -0.97 \\
    Artworks & -0.96 & -0.99 & -0.94 & -0.87 & -0.77 & -0.91 & -0.79 & -0.93  & -0.90 & -0.97 & -0.88 & -0.97    \\
    News  & -0.57 & -0.72 & -0.93 & -0.99 & -0.45 & -0.66 & -0.68 & -0.85  & -0.82 & -0.93 & -0.92 & -0.98  \\
    NYC  & -0.85 & -0.91 & -0.80 & -0.90 & -0.93 & -0.97 & -1.00 & -1.00  & -0.95 & -0.99 & -0.86 & -0.94  \\
    USA & -0.87 & -0.96 & \textemdash & \textemdash & -0.88 & -0.91 & \textemdash & \textemdash  & -0.71 & -0.85 & -0.74 & -0.89 \\
    World  & \underline{-0.84} & \underline{-0.93} & \underline{-0.89} &\underline{-0.92} & \underline{-0.77} & \underline{-0.86} & \underline{-0.83} & \underline{-0.78}  & \underline{-0.89} & \underline{-0.97} & \underline{-0.86} & \underline{-0.95}  \\  \midrule
    \multicolumn{13}{l}{- \textit{`K' and `Sp' denote the Kendall and Spearman rank correlation coefficients respectively.} }\\ \bottomrule
    \end{tabular}}
\end{table}

\subsection{Sensitivity experiments}

To assess the robustness of our findings with respect to key parameters—such as generation temperatures, the sliding window strategy for the probe model, and prompt selection—we conducted a series of sensitivity experiments, as presented below.  

\textbf{Generation temperature.} Temperature controls the creativity and randomness of LLM responses: lower values typically result in more deterministic outputs, while higher values lead to more varied and stochastic responses. In this experiment, we regenerate responses using different temperatures, ranging from 0.1 to 1.5. We then report the correlation coefficients under each temperature setting, as shown in the Figure \ref{fig:temperature}. The results show that the correlation remains relatively stable across the temperature range of $[0.1, 1]$, which covers the standard default range for most LLMs.

\textbf{Sliding window strategy.} The window and stride control how many samples are included in each segment and determine the total number of segments. Too few samples per segment can lead to unstable probe training, while too few segments may render the correlation estimates unreliable. In this experiment, we plot the Spearman correlation scores between response uncertainty (variance) and probe performance (Spearman rank coefficients) under different window and stride settings in Figure \ref{fig:sliding_window} on the \textit{Figures} dataset. Despite slight variations in the coefficient values, our main findings are not sensitive to this selection, as long as the two factors are not set to extreme values. 

\textbf{Prompt.} Different prompts inevitably affect the quality of generated responses, especially when advanced prompting strategies such as chain-of-thought reasoning or in-context examples are employed. In this study, we experiment with various simple prompts while keeping the core objective fixed: querying LLMs to answer a specific target concept. These variations include instructions such as asking the model to ``play as a historian'' or adding restriction for output format to guide the response. The prompt templates and results, shown in Table ~\ref{tab:prompts} of Appendix \ref{sec:sensitivity_appendix}, indicate that the correlation persists regardless of the slight difference in prompt template. 

\section{Exploring the mechanisms underlying the correlation}  
\label{sec:analysis}

\begin{figure}[htbp]
\begin{minipage}{0.68\textwidth}
  \centering
  \includegraphics[width=\linewidth]{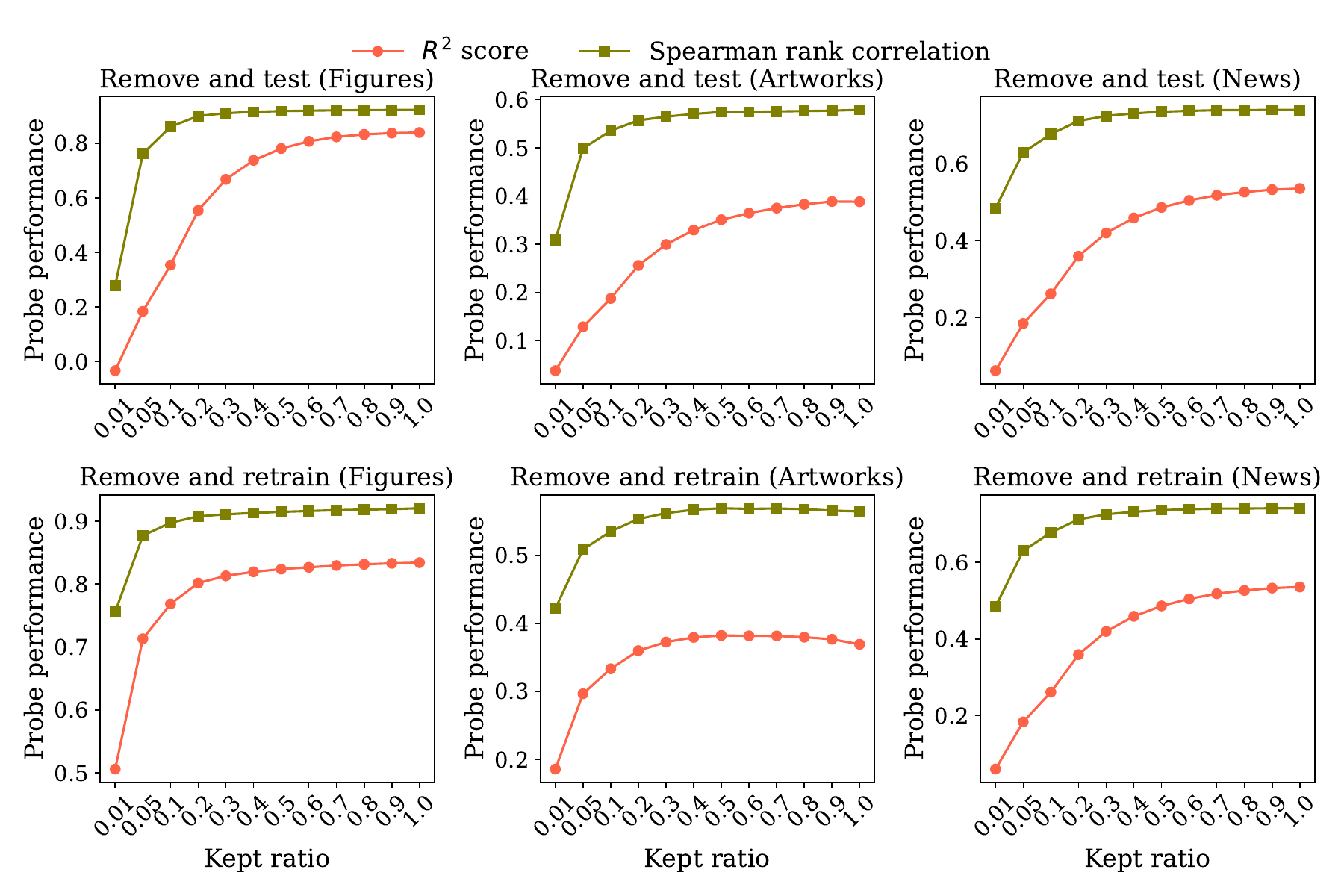}
  \caption{Llama 3.1 (8B) model. We gradually remove unimportant features to LLM responses and observe the probe performance drop.}
  \label{fig:fea_import_51}
\end{minipage}%
\hfill%
\begin{minipage}{0.28\textwidth}
  \centering
  \includegraphics[width=\linewidth]{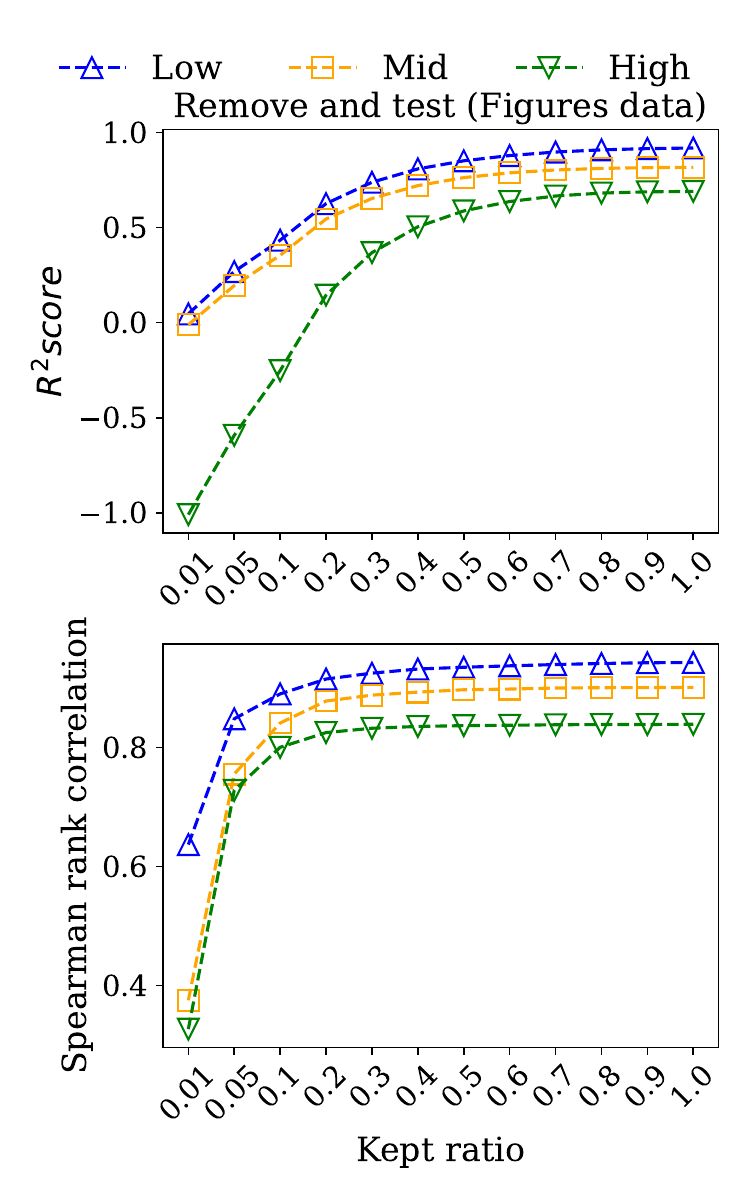}
  \caption{Remove and test on differently uncertain subsets, Llama 3.1 (8B).}
  \label{fig:fea_import_52}
\end{minipage}
\end{figure}

In this study, we analyze the correlation between response uncertainty and probe performance from a causal perspective, targeting to identify a confounding factor that could influence both. We hypothesize that \textit{the target concept is encoded in a set of features within the LLM embeddings that are shared by both the probe model and the generation process}. 

To investigate this, we first examine the importance scores of latent representation $\phi(x) \in \mathbb{R}^d$ with respect to an LLM’s response, using AttnLRP \cite{10.5555/3692070.3692076,arras2025close}. This yields a feature importance vector $\sigma(x) \in \mathbb{R}^d$, where each value reflects the contribution of the corresponding feature of $\phi(x)$ to the that response. A higher importance score indicates that removing the corresponding feature would result in a significant degradation in response quality. \textit{If both the probe model and the LLM generation rely on the same set of features, we would expect the probe model to preserve similar predictions when using the important features of response generation, compared to the original full representation. }

After obtaining the feature importance vector by tracing back from the response, we gradually mask the unimportant features by zeroing them out and observe the performance drop in the probe model.   We evaluate the degradation from two perspectives: (i) freezing the probe weights and directly test the probe performance on masked representations \cite{wang2024gradient}. Note this probe model is trained on the entire dataset; and (ii) masking the important features across all samples, retraining the probe model, and reporting its performance on the masked test set—following the Remove and Retrain (RoAR) paradigm \cite{adebayo2018sanity}. 

The results are reported in Figure~\ref{fig:fea_import_51}, where the $x$-axis represents the ratio of important features preserved (with the remaining features zeroed out), and the $y$-axis indicates the performance of the probe model. From them, we observe that even when only a small fraction of the most important features for the LLM response is preserved, the probe model can achieve performance comparable to using the full set of features. For example, preserving the top-30\% important features could achieve the similar results with the entire features on the Figure dataset in term of $R^2$ score. In the remove-and-retraining experiments, keeping top-20\% important features could achieve a good fit for probe models. These experiments provide strong evidence supporting the observed correlation: \textit{both the LLM and the probe model rely on the similar set of features relevant to the target concept}.

\subsection{Higher uncertainty implies more important features}

In this section, we present both theoretical and empirical analyses to investigate \textit{why response uncertainty is negatively correlated with probe performance across subsets with varying levels of uncertainty}.

\textbf{Mathematical analysis:} Suppose the set of important features for a given response $r_i$ is denoted as $\hat{\phi}_i(x) \in \mathbb{R}^k \subseteq \phi(x)$, where $k\ll d$, following the sparse representation findings of \cite{luo2024sparsing,voita2019analyzing}. If an LLM exhibits higher variation in its response to the same question, different responses are likely to track back to different important features in $\phi(x)$. This suggests that a larger number of features (i.e., the union of $\hat{\sigma}_i(x)$) should be deemed important, resulting in a more broadly distributed importance scores. According to the Lasso oracle inequality in compressed sensing domain \cite{buhlmann2011statistics}, the generalization bound of ridge regression is proportional to the ratio of important feature used with a constant probability. This implies that the probe is likely to perform worse when more features contribute significantly to the prediction, as it becomes harder to learn a stable and low-dimensional mapping. The detailed mathematical analysis is provided in the Appendix section \ref{apd:formal_analysis}. 

To verify this, we construct three subsets of 5,000 samples, selected based on varying levels of response uncertainty, referred to as Low, Mid, and High for brevity. We train a probe model for the entire dataset and observe the performance drop for each subset as unimportant features are progressively masked. The experimental results on the Figures dataset using Llama 3.1 (8B) are shown in Figure~\ref{fig:fea_import_52}. We can clearly see that the probe trained on the high-uncertainty subset requires a larger number of features to maintain comparable performance, whereas the probe trained on the low-uncertainty subset achieves similar performance using relatively fewer features. For example, preserving the top 40\% of important features approximately restores the probe performance for the high-uncertainty subset, while retaining only the top 20\% is sufficient for the low-uncertainty subset, as shown in the upper figure. In the bottom figure, preserving the top 5\% of important features is sufficient to achieve a similar Spearman score for the probe performance on the low-uncertainty subset, whereas 10\% is needed for the high-certainty subset. Due to page limit, we put more experiments that support this conclusion in the Appendix section \ref{apd:additonal_exp}. 

\subsection{Probe examples guided by response uncertainty}

\begin{figure}[htbp]
    \centering
    \begin{subfigure}{0.33\textwidth}
        \centering
        \includegraphics[width=\linewidth]{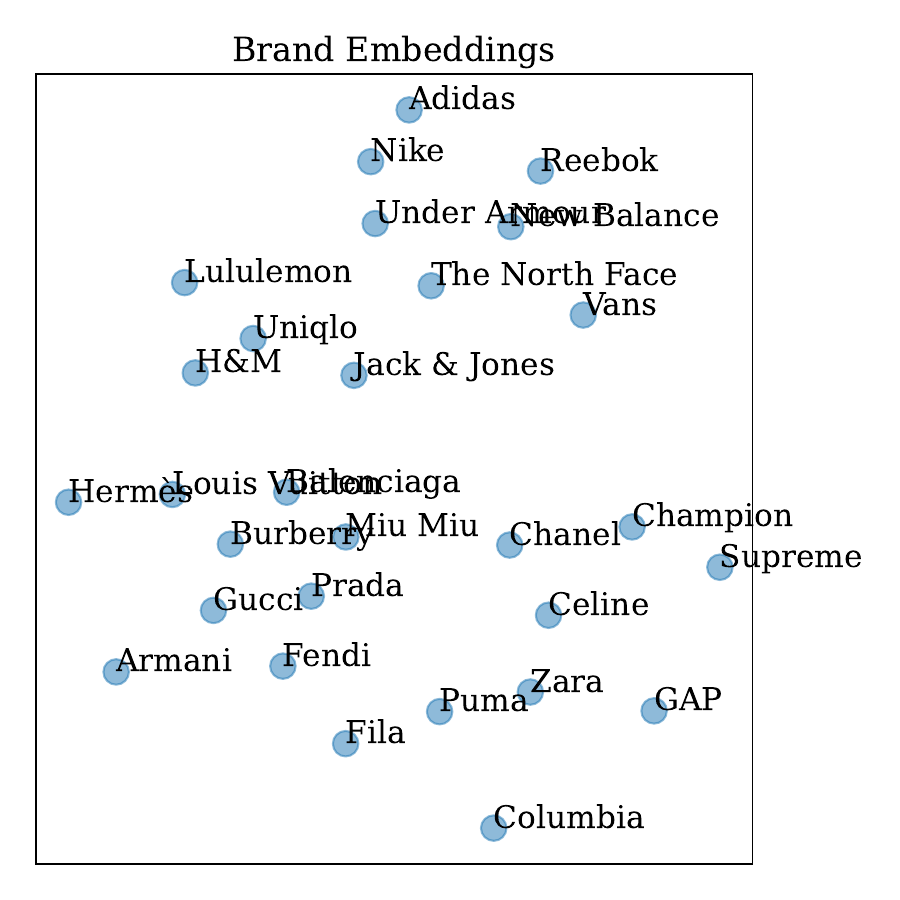}
        \caption{Brands}
        \label{fig:brands}
    \end{subfigure}%
    \hfill%
    \begin{subfigure}{0.33\textwidth}
        \centering
        \includegraphics[width=\linewidth]{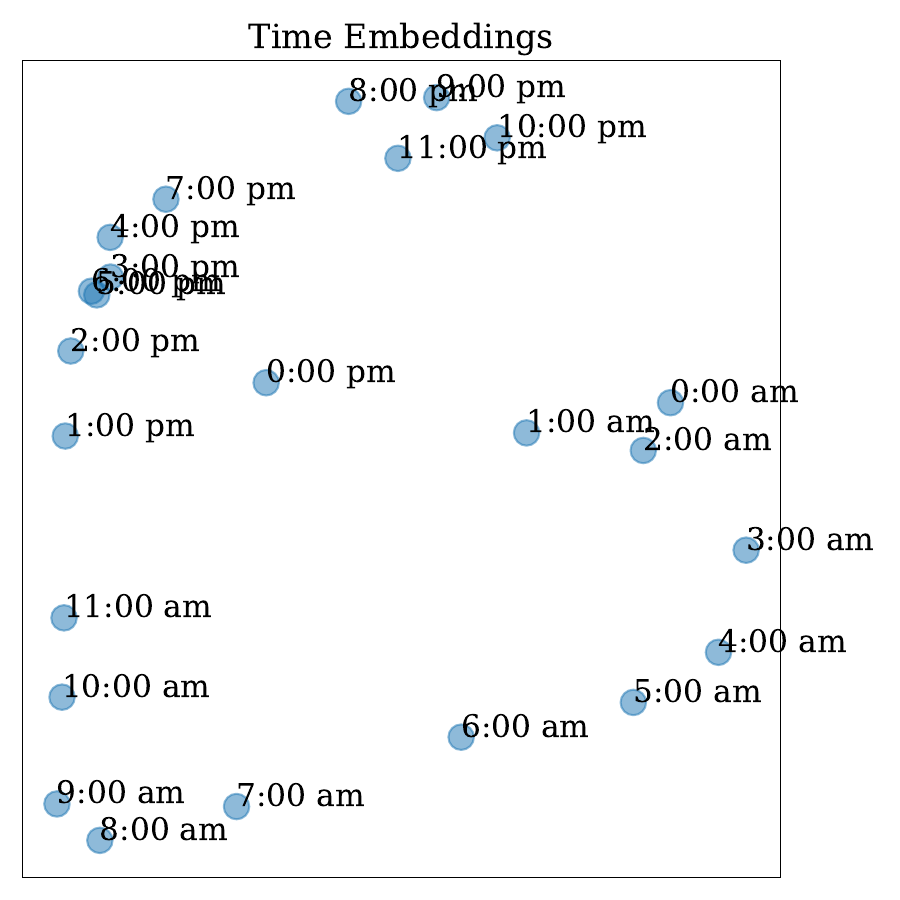}
        \caption{Time}
        \label{fig:time}
    \end{subfigure}%
    \hfill%
    \begin{subfigure}{0.33\textwidth}
        \centering
        \includegraphics[width=\linewidth]{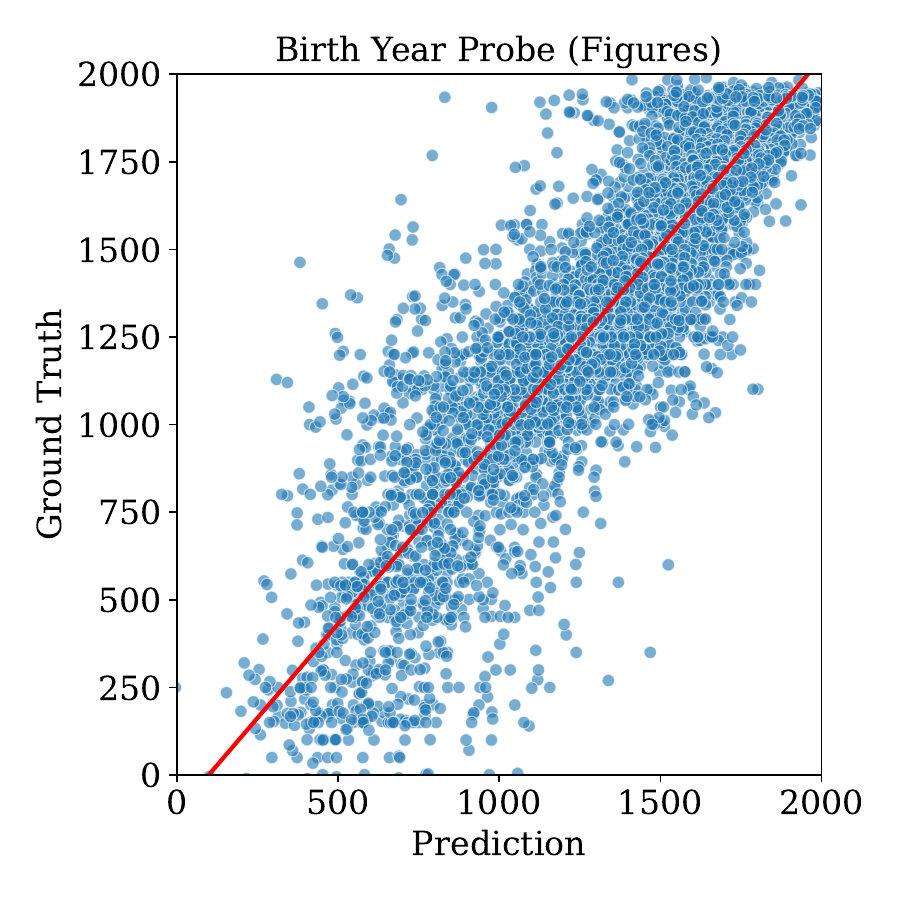}
        \caption{Birth Year}
        \label{fig:birth_year}
    \end{subfigure}
    \caption{Three commonsense examples where LLM embeddings align well with human knowledge or are easily probed, identified through consistent LLM responses.}
    \label{fig:examples}
\end{figure}

Guided by our hypothesis, we identified several examples in which LLM embeddings encode human-interpretable knowledge that can be effectively visualized or extracted using a probing model, shown in Figure \ref{fig:examples}. In these cases,  LLM also produces highly consistent responses to questions like ``What is the category of \{\}'' and ``What hour is after \{\}?'', as shown in Table \ref{tab:example_prompt} (Appendix). We extract the embedding of each entity name from layer 25 of Llama 3.1 (8B). For the first two examples, we project the embeddings into two dimensions using T-SNE. For the third example, we train a linear probe to predict the birth year from the same embeddings.

As shown in Figure~\ref{fig:examples}, the left panel visualizes embeddings of common brands such as Nike, Adidas, and Armani. We observe that luxury brands (e.g., LV, Burberry) cluster together, while sports brands (e.g., Nike, Adidas, Puma) form a separate cluster—indicating that the LLM captures categorical distinctions among brand types. The middle panel shows temporal representations from 0 a.m. to 11 p.m., revealing a circular pattern in the embedding space that aligns with typical daily activity cycles. In the right figure, we apply the linear regression probe to predict birth year from the embeddings of historical figures' names. We see that our probe achieve good fit compared to the results for death year prediction reported in \cite{gurnee2024language}. This example shows that the embedding of a figure's name encodes rich information, which can be effectively extracted using different probing methods. 

From the above three examples, we also observe that for mono-semantic entities, the  embedding visualization often aligns well with human prior knowledge. In contrast, for entities with stereo-semantic meanings, a probe model is required to extract a specific target concept. 

\section{Related work}

\textbf{LLM interpretability.} The lack of interpretability limits the deployment of LLMs in high-stakes applications. To address this challenge, a range of approaches has been proposed to explain and interpret the behaviors and predictions of LLMs \cite{10.1145/3639372,singh2024rethinking,wang2024gradient}. These methods include assessing the importance of input tokens in next-token prediction \cite{10.5555/3692070.3692076,arras2025close}, analyzing transformer circuits \cite{elhage2021mathematical,nanda2023progress}, and elucidating the reasoning processes behind LLM outputs \cite{wei2022chain,yao2023tree}. 

Our study falls within the scope of probe-based explanation methods, which aim to interpret LLMs at a global level. A probe—typically a simple, trained model—is utilized to detect the presence of a target concept from the embeddings produced by the LLM, usually adhering to the linear representation hypothesis \cite{mikolov-etal-2013-linguistic}. Generally, probe models are trained on external datasets containing training instances of the target concepts (e.g., low-level stripes in TCAV \cite{kim2018interpretability}, color attributes \cite{patel2022mapping}, or spatial and temporal concepts \cite{gurnee2024language}, game status \cite{li2023emergent}). Some other works propose to automatically learn concepts during CNN training without relying on external probe datasets \cite{koh2020concept,yeh2020completeness,burnsdiscovering}. A recent study provides strong evidence supporting circular representations for temporal concepts such as `week' and `year,' thus challenging the traditional linear hypothesis \cite{mikolov-etal-2013-linguistic}. However, to the best of our knowledge, no prior work has comprehensively investigated the relationship between probe performance and generative behavior in LLMs. 

\textbf{LLM hallucination.}  LLMs often generate non-factual or fabricated responses that deviate from grounded knowledge, stemming from various factors such as insufficient coverage in the training data or limited model capacity. For uncertainty that arises when the ground truth is deterministic—referred to as epistemic uncertainty \cite{abbasi2024believe}—researchers commonly use uncertainty measures to quantify the dispersion across multiple responses. Most studies \cite{kadavath2022language,farquhar2024detecting,kuhn2023semantic,kossen2025semantic} focus on developing metrics to more accurately measure response uncertainty from different perspectives, while largely overlooking the implications of these measurements.

\section{Limitations and discussion}
Although our extensive experiments demonstrate a strong correlation between probe performance and response uncertainty, several interesting phenomena remain poorly understood. We discuss these observations below and also highlight open questions for future work. 

\textbf{Probe weights and prompt embeddings.}  A growing body of evidence shows that LLM hidden representations encode conceptual information that can be effectively extracted using probing models. Simultaneously, such information can also be elicited through carefully designed prompts. We hypothesize that prompts function in a manner analogous to probes, although the mechanism remains poorly understood. In future work, it would be valuable to gather concrete evidence to test this hypothesis.  

\textbf{What happened during instruction tuning.} In our experiments, a stronger correspondence between probe performance and response consistency is observed only in instruction-tuned models. In contrast, for base LLMs, even when the probe achieves a good fit, the generated responses often remain disorganized and fail to answer questions accurately. This observation motivates further investigation into the parameter changes introduced by instruction tuning \cite{minder2025robustly}, particularly in the later layers of the model. 

\textbf{Concept encoding and decoding process.} Our case study demonstrates that LLMs encode rich information—for example, both birth and death years, within the representation of a person's name—which can be extracted using different probes. Similar conclusions have been drawn by \citet{adly2024monosemanticity}. We are particularly interested in understanding how such information is encoded and the internal mechanisms that govern its decoding, whether through probing models or user prompts.  

\textbf{A unified interpretability framework.} In this paper, we present strong evidence between these response uncertainty and LLM probe, suggesting that either method could be used to evaluate LLMs depending on practical availability. For example, response uncertainty may serve as a lightweight diagnostic tool when probing is not feasible. Alternatively, we could investigate whether a probe model applied to latent representations can be used to estimate response uncertainty \cite{kossen2025semantic}. More importantly, it is worth investigating whether similar correlations exist across other interpretability methods, with the long-term goal of developing a unified framework for LLM explanation. 

\section{Conclusion}
In this paper, we demonstrated a strong correlation between LLM response uncertainty and probe performance. Specifically, when the probe achieves higher performance, the LLM tends to produce more consistent responses, and vice versa. Furthermore, we delve deeper into this correlation from the feature importance analysis. Our findings reveal that high response uncertainty is typically associated with a larger number of important features, making it more challenging to effectively train a probing model. Finally, we identified concrete examples characterized by low response uncertainty, where the representation visualizations closely align with human knowledge or can be easily probed. 

\bibliographystyle{plainnat}
\bibliography{reference}

\newpage 

\appendix

\section{Dataset description}
\label{apd:dataset}

Our experiments are conducted on three spatial and three temporal datasets collected by \citet{gurnee2024language}. Each dataset contains the names of entities—such as celebrities, news headlines, artworks, or places—along with their associated spatial and temporal attributes. In \cite{gurnee2024language}, the authors propose learning a regression probe to extract spatial or temporal information from entity embeddings. Table~\ref{tab:dataset} summarizes the number of samples in each dataset and provides a representative example. Since we prompt LLMs to generate responses related to specific concepts, we also list the corresponding prompts. Here we use simple prompts only, without employing advanced prompting strategies such as chain-of-thought or in-context examples. By default, the temperature for response generation is set to 1, which is the default setting in the HuggingFace pipeline generator. 

\begin{table}[htbp]
    \centering
    \caption{Descriptions of the datasets and prompts used to generate LLM responses.}
    \label{tab:dataset}
     \scalebox{0.95}{
    \begin{tabular}{l l l l}
    \toprule
    Dataset & Count & Example & Prompt Template \\ \midrule
    Figures & 37,539 & ``Cleopatra'' & Return the death year of \{ \} \\
    Artworks & 31,321 & ``Stephen King’s It'' & Return the release year of \{ \} \\
    News & 28,389 & ``Dozens Rescued From ... '' & Return the publish year of \{ \} \\ 
    NYC & 19,838 & ``Borden Avenue Bridge'' & Return the latitude and longitude of \{\} in NYC\\ 
    USA  & 29,997 & ``Columbia University'' & Return the latitude and longitude of \{\} in USA \\
    World & 39,585 & ``Los Angeles'' & Return the latitude and longitude of \{\} \\ \bottomrule
\end{tabular}}
\end{table}

\section{Additional implementation details}
\label{apd:implementation}

In this section, we provide additional details about our experimental setup. Table~\ref{tab:setting} summarizes the settings used for training the probe model. Due to the higher feature dimensionality in large LLMs such as LLama 3.1 (70B), a larger number of samples is required. To manage this, we set different window sizes and strides. For each sample, we extract the representation from the 25th layer, using the hidden state of the last token. 

During probe training, we randomly split the data—whether a segment, a subset, or the entire dataset depending on the experimental setting—into training and test sets at a 
$4:1$ ratio. The training set is further divided into two parts: one for training the probe and the other for tuning the trade-off parameter $\lambda$. We report both the $R^2$ score and the Spearman rank correlation coefficient on the test set. 

By default, we use the prompt template shown in Table~\ref{tab:dataset} to generate multiple responses for each sample. We also experiment with alternative prompt formats on the Figures dataset with Llama 3.1 8B model, as listed in Table~\ref{tab:prompts}. 

\begin{table}[htbp]
    \centering
    \caption{Target layer, representation dimension and sliding window settings (window size and stride) for different LLMs.}
    \label{tab:setting}
     \scalebox{0.95}{
    \begin{tabular}{l l l l l l}
    \toprule
    Model & Target layer & \#Layers & Dimension & Window size & Stride \\ \midrule
    Llama 3.1 (8B) & 25 &32 & 4096 & 6000 & 1000\\
    Llama 3.1 (70B) & 60 &80 &8192 & 12000 & 1000\\
    Qwen 2.5 (14B) & 36 &48 &5120 & 8000 & 1000\\ 
    Qwen 2.5 (72B) & 60 & 80&8192 & 12000 & 1000\\ 
    Mistral-Small (24B) & 30 &40 & 5120 & 8000  & 1000\\
    Gemma 2 (27B) & 35 & 48&4608 & 8000 & 1000 \\ \bottomrule
\end{tabular}}
\end{table}

\section{Entropy as the uncertainty estimator}
\label{apd:entropy_estimator}

Entropy is a widely used estimator for measuring uncertainty over multiple responses in prior studies \cite{farquhar2024detecting,kuhn2023semantic,kossen2025semantic}. However, entropy is not applicable to continuous values such as longitudes and latitudes in spatial datasets. In Table~\ref{tab:entropy_uncertainty}, we adopt entropy as the uncertainty estimator to measure variation in temporal responses. Figure \ref{fig:entropy_trend} includes the trend between response uncertainty and probe performance across segments. Experimental results show that our conclusions still hold under this metric. 

\begin{table}[htbp]
    \centering
    \caption{Correlation between probe performance and LLM response uncertainty on Llama 3.1 (8B, 70B), Qwen 2.5 (14B, 72B), Mistral-Small (24B), and Gemma-2 (27B) across three temporal datasets.}
    \label{tab:entropy_uncertainty}
    \scalebox{0.9}{
    \begin{tabular}{l| cc | cc | cc | cc | cc | cc  cc}
    \toprule
    \multirow{2}{*}{Dataset} & \multicolumn{2}{c|}{LLama 8B} & \multicolumn{2}{c|}{LLama 70B} & \multicolumn{2}{c|}{Qwen 14B} & \multicolumn{2}{c|}{Qwen 72B} & \multicolumn{2}{c|}{Mistral 24B} & \multicolumn{2}{c}{Gemma 27B} \\ \cmidrule{2-13}
     & K & Sp & K & Sp & K & Sp &  K & Sp & K & Sp & K & Sp \\ \midrule
    \multicolumn{13}{c}{\textit{Correlation coefficient: response uncertainty (\textbf{entropy}) vs probe performance ($R^2$)}. } \\ \midrule
    Figures  & -0.96 & -0.87 & -0.90 & -0.77 & -0.82 & -0.76 & -0.96 &-0.85   & -0.97 & -0.89 & -0.99 &-0.95  \\
    Artworks &-0.99  & -0.98 & -0.99 & -0.96 & -0.97 &-0.90  & -0.87 &-0.71   & -0.98 & -0.91 & -0.88 &-0.76   \\
    News  & -0.99 & -0.97 & -0.99 & -0.98 & -0.94 & -0.88 & -0.91 & -0.81  & -0.96 & -0.90 & -0.98 &-0.90 \\ \midrule
    \multicolumn{13}{c}{\textit{Correlation coefficient: response uncertainty (\textbf{entropy}) vs probe performance (\textbf{Spearman rank score})}.} \\ \midrule
    Figures  & -0.98 & -0.99 & -0.90 & -0.97 & -0.90 & -0.83 & -0.80 &-0.93   & -0.91 & -0.97 & -0.96 &-0.99 \\
    Artworks &-0.97  & -0.99 & -0.98 & -0.99 & -0.88 & -0.96 & -0.80 & -0.91  & -0.91 &-0.98  & -0.90 &-0.97    \\
    News  & -0.95 & -0.98 & -0.98 &-0.99  & -0.89 & -0.95 & -0.81 & -0.90  & -0.96 & -0.83 & -0.93 &  \\  \midrule
    \multicolumn{13}{l}{- \textit{`K' and `Sp' denote the Kendall and Spearman rank correlation coefficients respectively.} }\\ \bottomrule
    \end{tabular}}
\end{table}

\section{The linearity of correlation} 
\label{apd:linear}

One interesting question is whether the correlation between response uncertainty and probe performance is linear. To investigate this, we report the Pearson correlation coefficients on three temporal datasets, as shown in Table~\ref{tab:pearson}. We include both variance and entropy as uncertainty estimators in our analysis.  Recognizing that different uncertainty estimators operate on different scales—and that probe performance itself is unaffected by the choice of uncertainty measure—we observe that the linearity of the correlation depends on the specific uncertainty estimator used.  This observation may also explain why \citet{kossen2025semantic} could train a linear model to predict uncertainty (semantic entropy) from hidden representations.

\begin{table}[htbp]
    \centering
    \caption{Pearson coefficient between probe performance and LLM response uncertainty on Llama 3.1 (8B, 70B), Qwen 2.5 (14B, 72B), Mistral-Small (24B), and Gemma-2 (27B) across three temporal datasets.}
    \label{tab:pearson}
    \scalebox{0.9}{
    \begin{tabular}{l| c | c | c | c | c | c  c}
    \toprule
    {Dataset} & {LLama 8B} & {LLama 70B} & {Qwen 14B} & {Qwen 72B} & {Mistral 24B} &{Gemma 27B} \\ \midrule
    \multicolumn{7}{c}{\textit{Pearson coefficient: \textbf{variance} vs \textbf{probe performance ($R^2$)}}. } \\ \midrule
    Figures  &  -0.46& -0.53 &-0.20 &-0.37 & -0.48 & -0.73  \\
    Artworks   & -0.36 & -0.54 & -0.20 &-0.52  &-0.50  &-0.30    \\
    News    &-0.07  &-0.24  &-0.3  & -0.15 & -0.54 & -0.40    \\ \midrule
    \multicolumn{7}{c}{\textit{Pearson coefficient: \textbf{variance} vs \textbf{probe performance (Spearman rank score)}}} \\ \midrule
    Figures  & -0.40 & -0.56 & -0.36 & -0.26 & -0.61 & -0.42  \\
    Artworks &-0.40  & -0.57 & -0.15 & -0.50 & -0.50 & -0.25      \\
    News  & -0.06 & -0.32 & -0.30 & -0.26 & -0.61 & -0.42    \\  \bottomrule
     \multicolumn{7}{c}{\textit{Pearson coefficient: \textbf{entropy} vs \textbf{probe performance ($R^2$)}}. } \\ \midrule
    Figures  &-0.93  & -0.96 & -0.97 & -0.96 &  -0.95& -0.96   \\
    Artworks & -0.98 &-0.98  & -0.98 & -0.89 & -0.94 & -0.91   \\
    News  & -0.99 & -0.99 & -0.99 & -0.93 & -0.97 & -0.94    \\ \midrule
    \multicolumn{7}{c}{\textit{Pearson coefficient: \textbf{entropy} vs \textbf{probe performance (Spearman rank score)}}} \\ \midrule
    Figures  & -0.96 & -0.97 & -0.81 &  -0.93& -0.97 &-0.97   \\
    Artworks &-0.98  & -0.99 & -0.97 & -0.89 &  -0.93&  -0.96     \\
    News  & -0.99 & -0.99 & -0.99 & -0.95 & -0.97 & -0.92   \\  \bottomrule
    \end{tabular}}
\end{table}

\section{Additional sensitivity experiments}
\label{sec:sensitivity_appendix}

\begin{figure}[htbp]
    \centering 
    \includegraphics[width=0.9\linewidth]{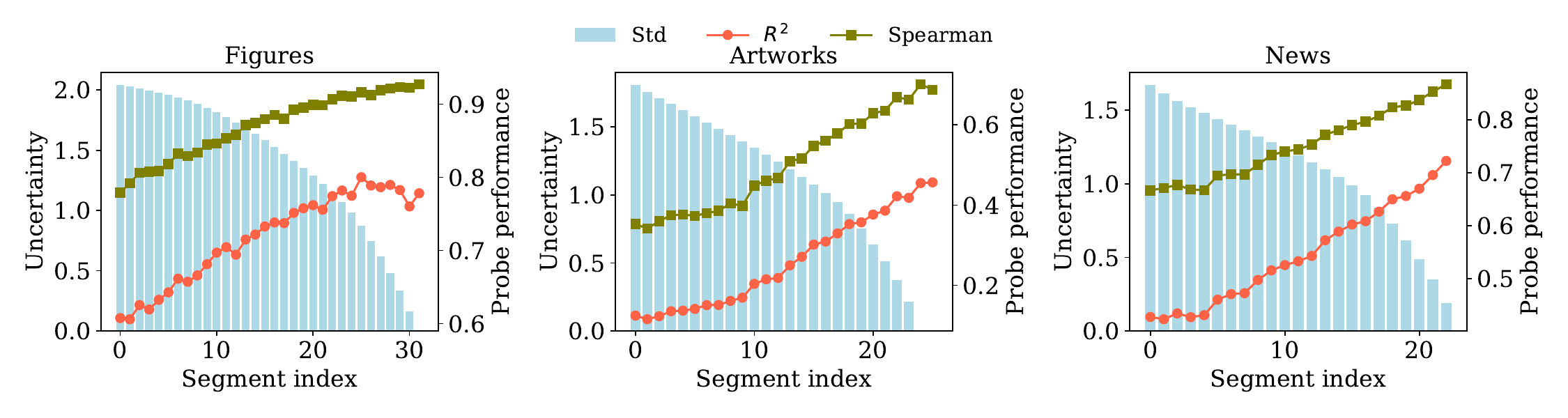}
    \caption{Trend between response uncertainty (entropy) and probe performance on the Llama 3.1 (8B) model over three temporal datasets.}
    \label{fig:entropy_trend}
      \includegraphics[width=0.9\linewidth]{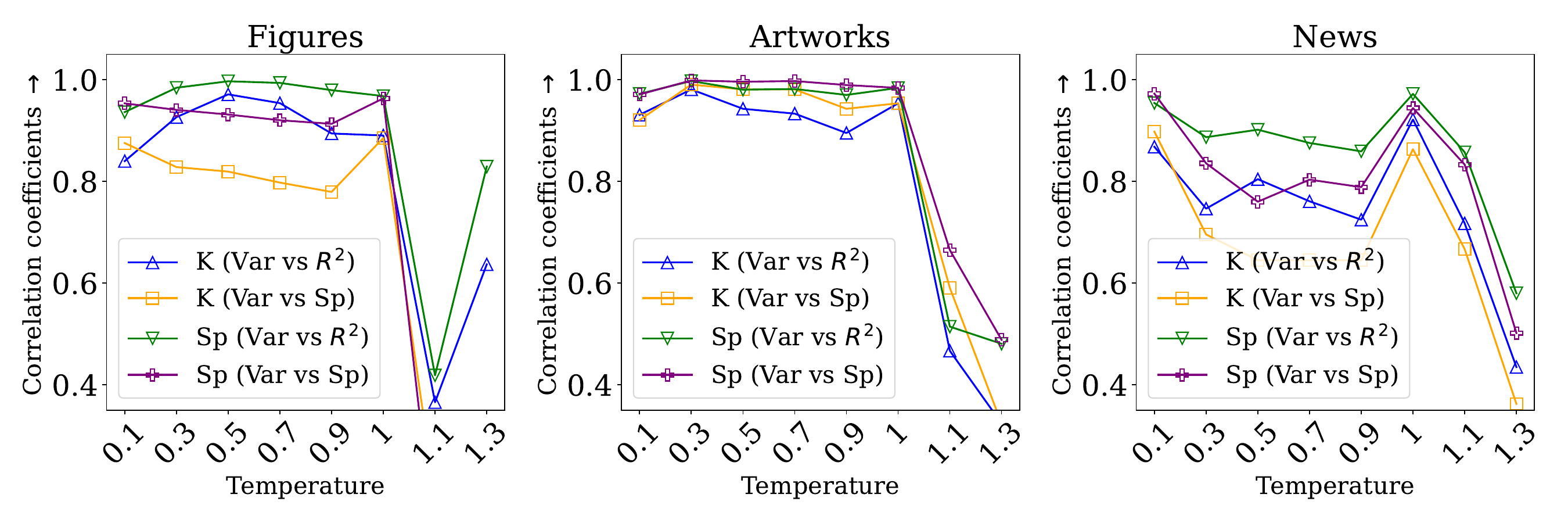}
    \caption{The correlation analysis with different temperatures in generation for Mistral-Small (24B). `Var', `$K$', and `Sp' denotes variance (uncertainty estimator), Kendall and Spearman rank correlation coefficients. Here we use absolute coefficient value for good visualization performance.}
    \label{fig:temperature_mistral}
\end{figure}

\begin{figure}[htbp]
\centering
     \begin{subfigure}{0.33\textwidth}
        \centering
        \includegraphics[width=\linewidth]{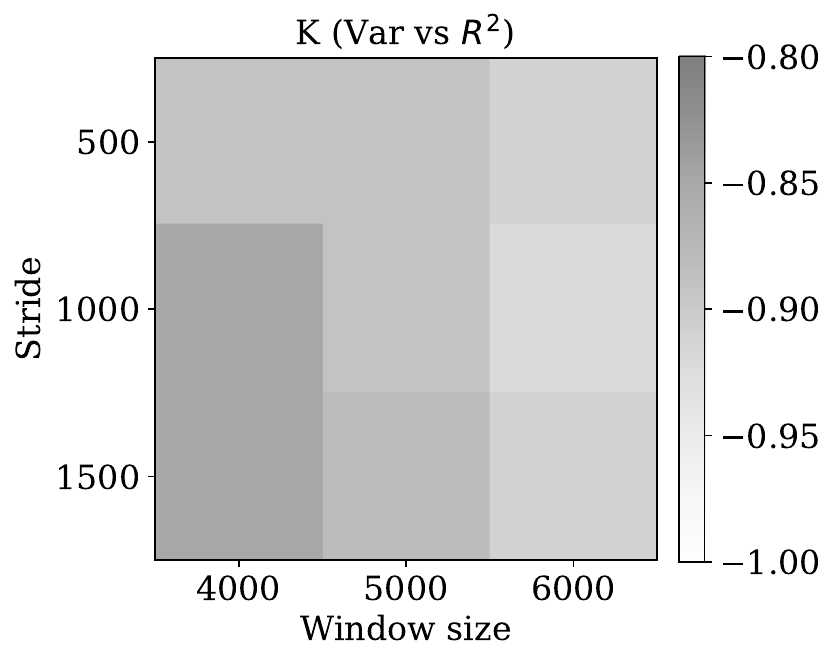}
        \label{fig:heat1}
    \end{subfigure}%
    \hfill%
    \begin{subfigure}{0.33\textwidth}
        \centering
        \includegraphics[width=\linewidth]{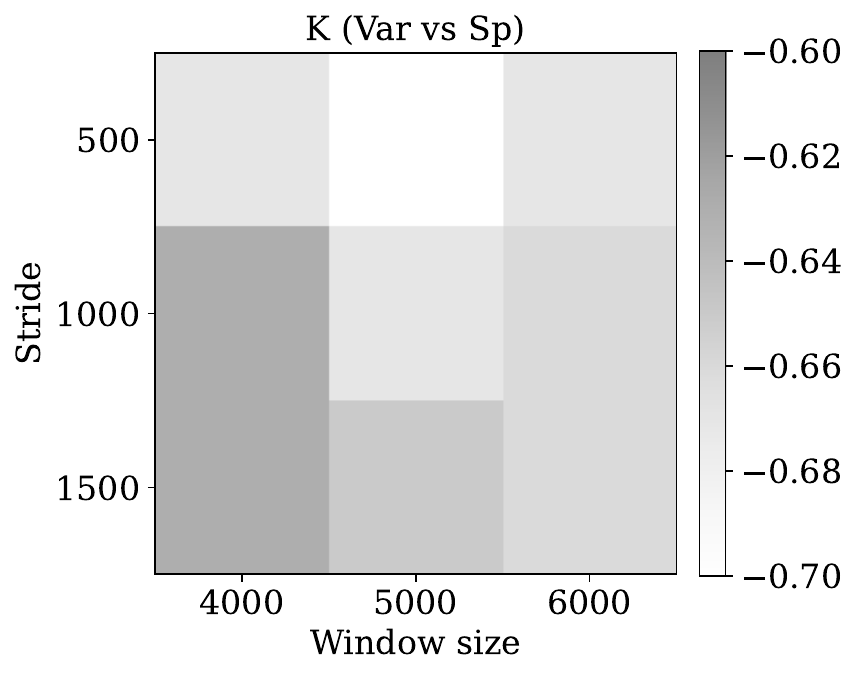}
        \label{fig:heat2}
    \end{subfigure}%
    \hfill%
    \begin{subfigure}{0.33\textwidth}
        \centering
        \includegraphics[width=\linewidth]{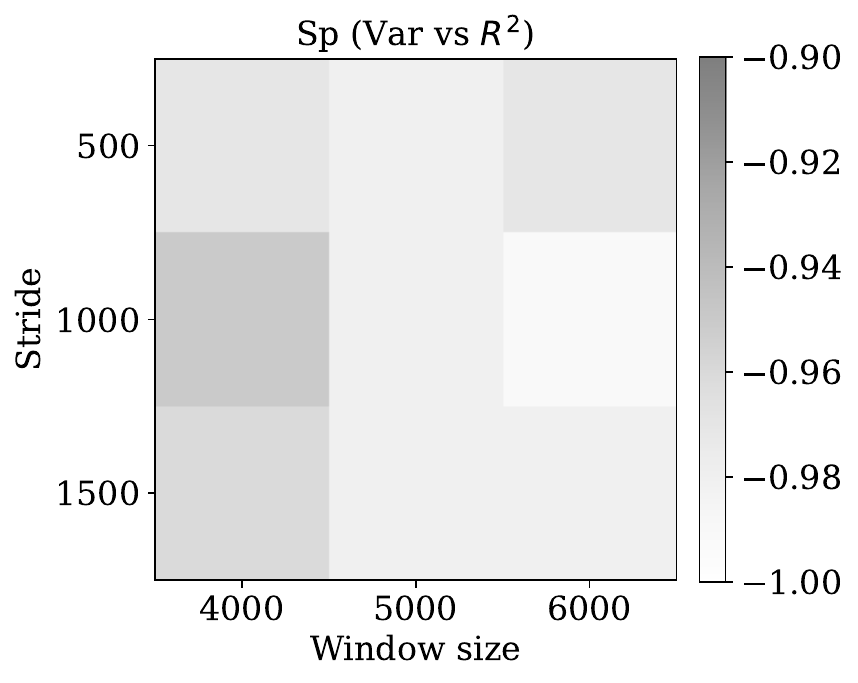}
        \label{fig:heat3}
    \end{subfigure}
    \caption{Correlation coefficient vs sliding window parameters, Llama 3.1 (8B) on Figures. `Var', `$K$', and `Sp' denotes variance (uncertainty estimator), Kendall and Spearman rank correlation coefficients.}
    \label{fig:all_heatmap} 
\end{figure}

In this section, we present additional experimental results related to generation temperature (Figure \ref{fig:temperature_mistral}), the sliding window strategy (Figure \ref{fig:all_heatmap}), and prompt templates \ref{tab:prompts} used for the probe model. Our conclusions hold across general settings, except in extreme cases—such as very high or low temperatures, overly small window sizes or large strides, or the use of complex prompts. 

\begin{table}[htbp]
    \centering
    \caption{The role of prompt template in correlation coefficient analysis on Llama 3.1 (8B) using the Figures dataset. `K' and `Sp' denotes Kendall and Spearman rank correlation coefficients, respectively.}
    \label{tab:prompts}
     \scalebox{1}{
    \begin{tabular}{p{7cm}|cc |cc}
    \toprule 
    \multirow{2}{*}{Prompt} & \multicolumn{2}{c|}{Variance vs $R^2$} & \multicolumn{2}{c}{Variance vs Sp}  \\ \cmidrule{2-5}
     & K & Sp & K & Sp\\ \midrule
    System prompt: Play as a historian and return the death year
of given person. User prompt: when did \{\} die? &-0.85 & -0.96 & -0.49& -0.66\\ \midrule
    System prompt: Return the death year
of given person. User prompt: when did \{\} die?  & -0.75 & -0.85& -0.43& -0.66\\ \midrule
    Return the death year of \{\}:  & -0.89 & -0.98& -0.67 & -0.82\\ \midrule
    What is the death year of \{\}? only return the year. & -0.61 & -0.78&-0.62&-0.76 \\ \bottomrule
\end{tabular}}
\end{table}

\section{Additional experiments of Section 5}
\label{apd:additonal_exp}
\begin{figure}[htbp]
\begin{minipage}{0.68\textwidth}
  \centering
  \includegraphics[width=\linewidth]{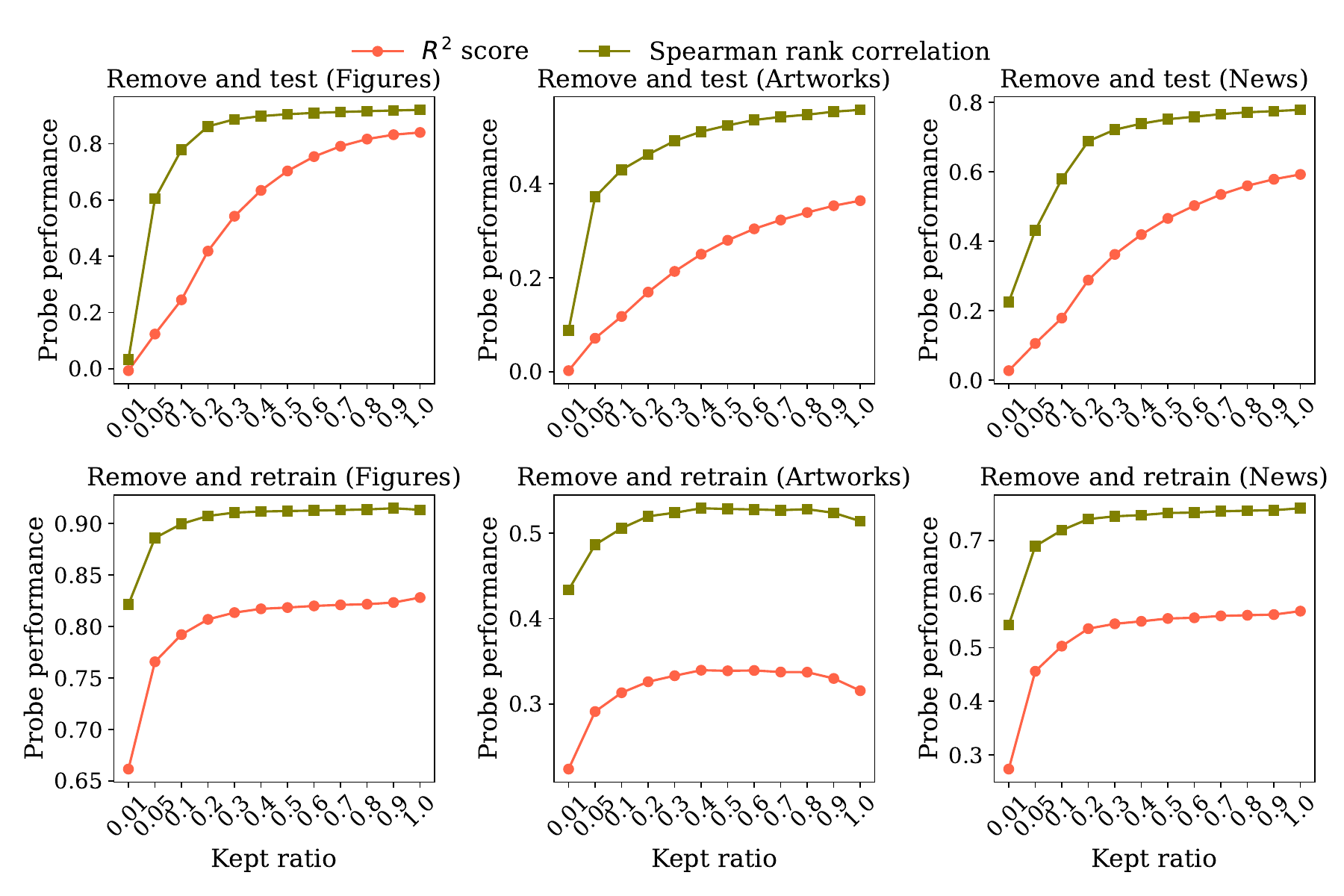}
  \caption{Llama 2 (13B) model. We gradually remove unimportant features to LLM responses and observe the probe performance drop.}
  \label{fig:fea_import_adp}
\end{minipage}%
\hfill%
\begin{minipage}{0.28\textwidth}
  \centering
  \includegraphics[width=\linewidth]{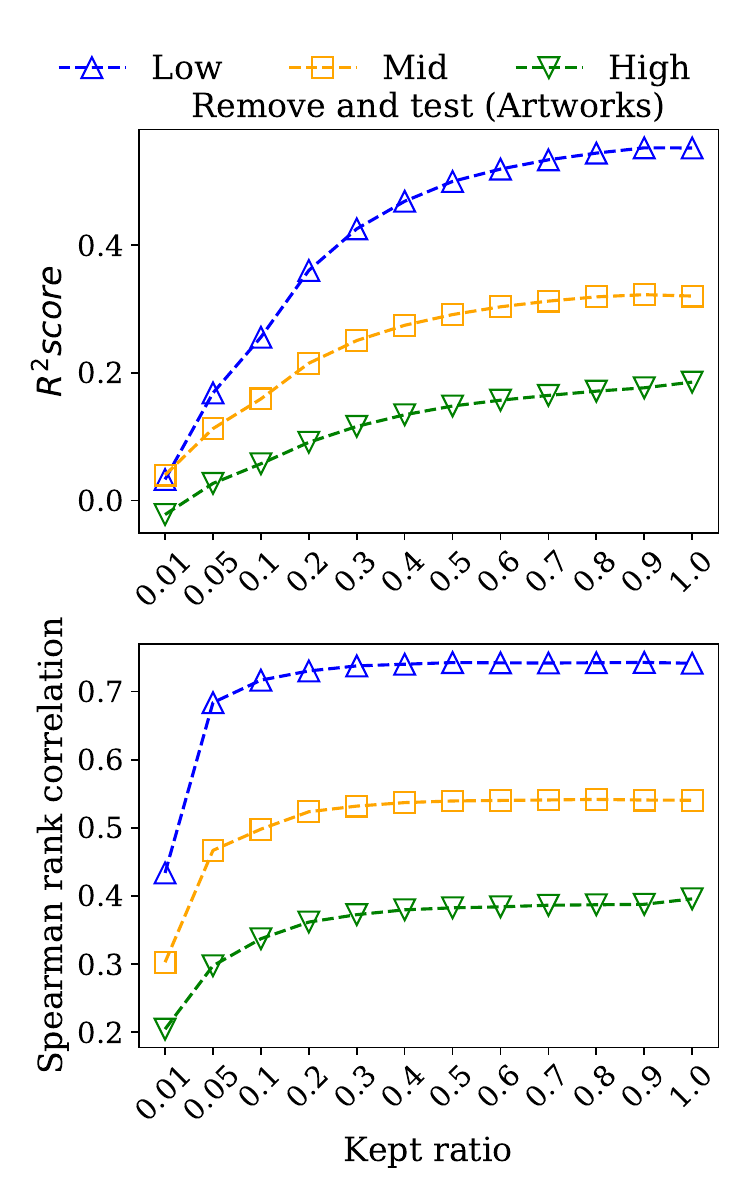}
  \caption{Remove and test on differently uncertain subsets, Llama 3.1 (8B).}
  \label{fig:fea_import_adp2}
\end{minipage}
\end{figure}

In this section, we present additional experiments related to Section~\ref{sec:analysis}. Since AttnLRP is specifically designed for the smaller Llama models, and does not support Llama 3.1 (70B), we use Llama 2 (14B) to study whether the probe model and the generation process rely on similar features in the input representation $\phi(x)$. The experimental results are shown in Figure~\ref{fig:fea_import_adp}. In the remove-and-retrain experiments, retaining the top 20\% of important features is sufficient to achieve a good fit for the probe models.

\begin{figure}
    \centering
    \includegraphics[width=\linewidth]{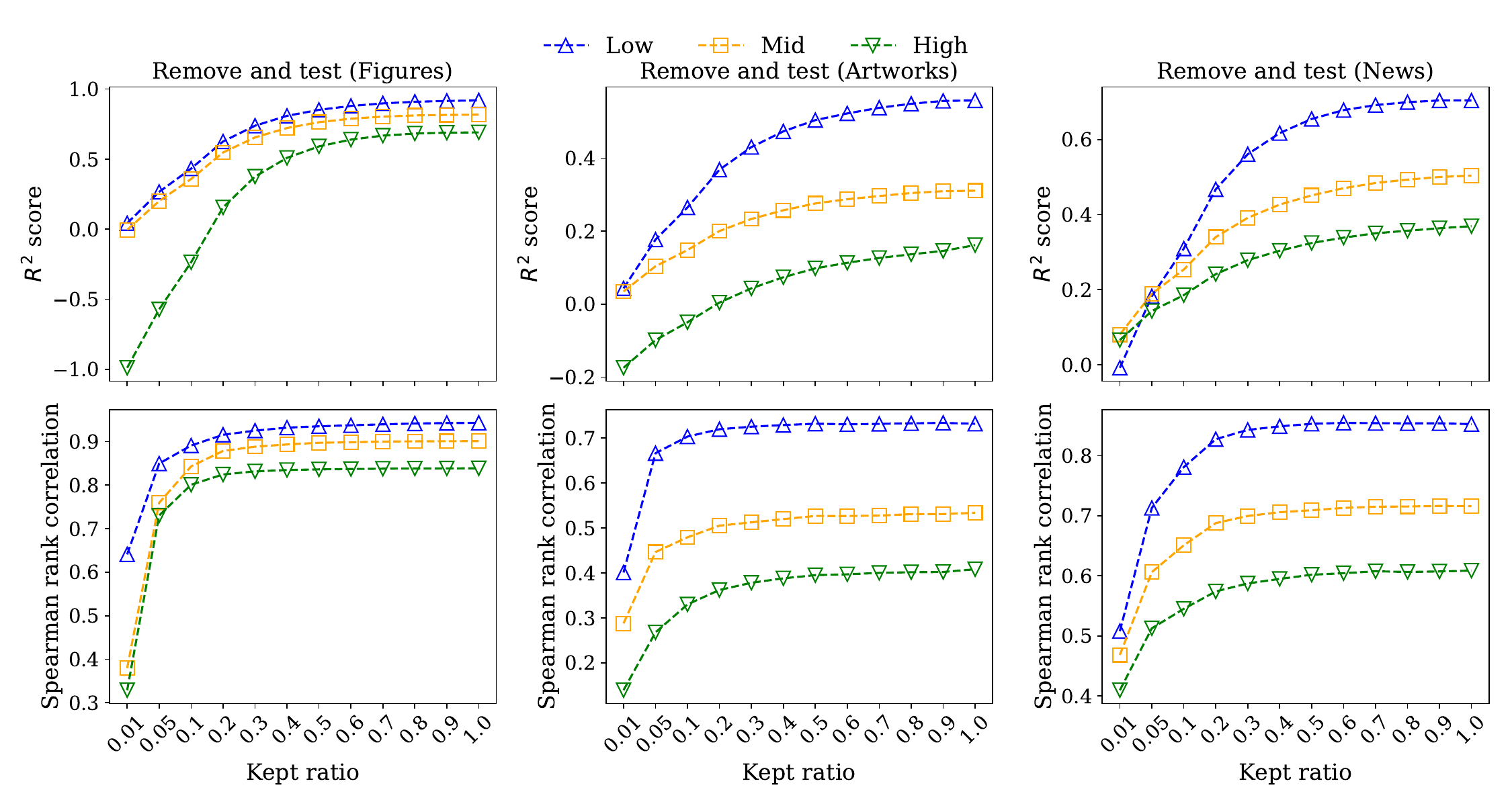}
    \caption{Remove and test on three subsets of varying uncertainty (measured by entropy), on Llama 3.1 (8B).  Probe training on high-uncertainty datasets requires more features to achieve higher results.}
    \label{fig:fea_import_entropy}
\end{figure}

Figure \ref{fig:fea_import_adp2} reports the remove and test results on different uncertainty subsets using Artworks dataset. From the bottom figure, we can see that preserving the top 5\% of important features approximately achieves the similar Spearman rank correlation for low-uncertainty subset while 30\% is needed for high-certainty subset. This experiment provides additional evidence to support our hypothesis. The same conclusion holds when using entropy as the uncertainty estimator to select three subsets with varying levels of uncertainty, as shown in Figure~\ref{fig:fea_import_entropy}. 

\section{Formal analysis of section 5.2}
\label{apd:formal_analysis}

Let us revisit the objective of investigating the underlying reasoning behind the negative correlation between response uncertainty and probe performance.  Given an sample $x$, we obtain the embedding of the target layer $l$, denoted as $\phi(x)$ where the layer index is omitted for simplicity. The final response is generated by forwarding $\phi(x)$ through the remaining layers, represented by $\varphi(\phi(x))$. Due to the stochastic nature of sampling during generation, the LLM may produce different responses for the same question $x$, denoted by $\{r_i,...,r_m\}$. With the help of AttnLRP \cite{10.5555/3692070.3692076}, we compute feature importance scores $\sigma_i(x)$ with respect to a single response $r_i$, where a higher value indicates that the absence of the corresponding feature would significantly affect the final output. 

Recent studies \cite{luo2024sparsing,voita2019analyzing} have shown that only a small subset of features is highly important for model predictions. Consistently, our experiments demonstrate that preserving the top 20\% of features yields competitive probing performance in Figure \ref{fig:fea_import_51}. Let $\hat{\phi}_i(x) \in \mathbb{R}^k \subseteq \sigma(x)$ denote the core set of important features in $\phi(x)$ with respect to response $r_i$, where $k \ll d$. If the model produces varying responses, the corresponding set $\hat{\phi}_i(x)$ are likely to differ, suggesting that a large number of features in $\phi(x)$ contribute to the final prediction. For example, if an LLM produces two different responses for the death year of George Washington—e.g., 1799 and 1796—different features in $\phi(x)$ will contribute to the final token position of these responses. 

Next we dive into the lower probe performance of high uncertainty response from Lasso oracle inequality in compressed sensing domain \cite{buhlmann2011statistics}. For a linear regression model with an $s$-sparse true parameters, we have 
\begin{equation}
\Pr\!\Bigl[
      L(\widehat{w})-\widehat{L}(\widehat{w})
      \;\le\;
      C\,\sigma^{2}\,\sqrt{\frac{s\log d}{n}}
\Bigr]
\;\;\ge\;1-\delta .
\label{eq:bound}
\end{equation} 
where, 
\begin{description}[leftmargin=5em, labelindent=2em]
  \item[$L(\widehat{w})$]:  Population (true) risk.  
  \item[$\widehat{L}(\widehat{w})$]:  Empirical risk on \(n\) samples.  
  \item[$d$]:  Total number of available features.  
  \item[$s\ll d$]:  Number of \emph{truly useful} (non-zero) features (sparsity).  
  \item[$\sigma^{2}$]:  Noise variance / loss Lipschitz constant.  
  \item[$C$]:  Universal numerical constant.  
  \item[$\delta$]:  Confidence level of the high-probability bound.  
\end{description}
With probability at least $1-\delta$, the gap between the empirical risk and the population (true) risk is upper-bounded by a sparsity-controlled term.
For an LLM with higher response uncertainty, a larger number of features contribute to the diverse responses compared to a model with lower uncertainty, making it more difficult to achieve accurate predictions with a linear regression probe. 

\section{Prompts in Section 5.3}

We list the prompts used to query the LLM for two examples from Section 5.3 in Table~\ref{tab:example_prompt}. In both cases, the LLM produced identical responses, exhibiting very low uncertainty. 

\begin{table}[]
    \centering
     \caption{Prompts and response uncertainty for examples in Section 5.3.}
    \begin{tabular}{p{7cm}|l} 
    \toprule
    Prompt & Entropy (Llama 3.1 (8B)) \\ \toprule
    What is the category of \{brand\}? Only select one from [Luxury, Fashion, Sports, Casual]   & 0.12 \\ \midrule
    What hour is after \{hour\}?   &  0 \\ \bottomrule
    \end{tabular}
   
    \label{tab:example_prompt}
\end{table}

\end{document}